\documentclass[runningheads]{llncs}
\usepackage{graphicx}
\usepackage{amsmath,amssymb}
\usepackage{color}
\usepackage{booktabs}
\usepackage{import}
\usepackage{tabularx}
\newcolumntype{Y}{>{\centering\arraybackslash}X}

\def\ie{\emph{i.e.}} 
 
\def\etal{\emph{et al.~}}
\newcommand*\samethanks[1][\value{footnote}]{\footnotemark[#1]}
\begin{document}

\title{Single Shot Scene Text Retrieval} 

\titlerunning{Single Shot Scene Text Retrieval}

\author{Llu{\'i}s G{\'o}mez\thanks{These authors contributed equally to this work} \and
Andr{\'e}s Mafla\samethanks \and
Mar\c{c}al Rusi\~{n}ol \and
Dimosthenis Karatzas}

\authorrunning{L. G{\'o}mez, A. Mafla, M. Rusi\~{n}ol and D. Karatzas}

\institute{Computer Vision Center, Universitat Aut\`{o}noma de Barcelona\\
Edifici O, 08193 Bellaterra (Barcelona), Spain. \\
\email{\{lgomez,andres.mafla,marcal,dimos\}@cvc.uab.es}}
\maketitle

\begin{abstract}
Textual information found in scene images provides high level semantic information about the image and its context and it can be leveraged for better scene understanding. In this paper we address the problem of scene text retrieval: given a text query, the system must return all images containing the queried text. The novelty of the proposed model consists in the usage of a single shot CNN architecture that predicts at the same time bounding boxes and a compact text representation of the words in them. In this way, the text based image retrieval task can be casted as a simple nearest neighbor search of the query text representation over the outputs of the CNN over the entire image database. Our experiments demonstrate that the proposed architecture outperforms previous state-of-the-art while it offers a significant increase in processing speed.
\end{abstract}

\section{Introduction}
\label{sec:intro}
The world we have created is full of written information. A large percentage of everyday scene images contain text, especially in urban scenarios ~\cite{lin2014microsoft,veit2016coco}.
Text detection, text recognition and word spotting are important research topics which have witnessed a rapid evolution during the past few years.
Despite significant advances achieved, propelled by the emergence of deep learning techniques~\cite{lecun2015deep}, scene text understanding in unconstrained conditions remains an open problem attracting an increasing interest from the Computer Vision research community.
Apart from the scientific interest, a key motivation comes by the plethora of potential applications enabled by automated scene text understanding, such as improved scene-text based image search, image geo-localization, human-computer interaction, assisted reading for the visually-impaired, robot navigation and industrial automation to mention just a few.

\begin{figure*}[t]
\centering
\includegraphics[width=1\linewidth]{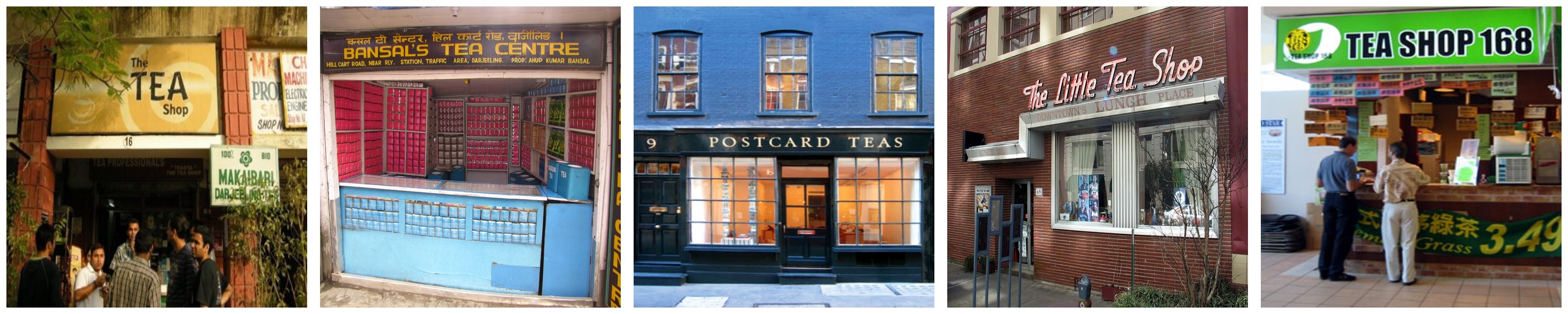}
\caption{The visual appearance of different tea shops' images can be extremely variable. It seems impossible to correctly label them without reading the text within them. Our scene text retrieval method returns all the images shown here within the top-10 ranked results among more than $10,000$ distractors for the text query ``tea''.}
\label{fig:tea_shops}
\end{figure*}

The textual content of scene images carries high level semantics in the form of explicit, non-trivial data, which is typically not possible to obtain from analyzing the visual information of the image alone.
For example, it is very challenging, even for humans, to automatically label images such as the ones illustrated in Figure~\ref{fig:tea_shops} as tea shops solely by their visual appearance, without actually reading the storefront signs.
Recent research actually demonstrated that a shop classifier ends up automatically learning to interpret textual information, as this is the only way to distinguish between businesses~\cite{movshovitz2015ontological}.
In recent years, several attempts to take advantage of text contained in images have been proposed not only to achieve fine-grained image classification~\cite{karaoglu2017text,bai2017integrating} but to facilitate image retrieval.
 
Mishra~\etal\cite{mishra2013image} introduced the task of scene text retrieval, where, given a text query, the system must return all images that are likely to contain such text. Successfully tackling such a task entails fast word-spotting methods, able to generalize well to out-of-dictionary queries never seen before during training.

A possible approach to implement scene text retrieval is to use an end-to-end reading system and simply look for the occurrences of the query word within its outputs. It has been shown~\cite{mishra2013image} that such attempts generally yield low performance for various reasons. First, it is worth noting that end-to-end reading systems are evaluated on a different task, and optimized on different metrics, opting for high precision, and more often than not making use of explicit information about each of the images (for example, short dictionaries given for each image). In contrary, in a retrieval system, a higher number of detections can be beneficial. Secondly, end-to-end systems are generally slow in processing images, which hinders their use in real-time scenarios or for indexing large-scale collections.

In this paper we propose a real-time, high-performance word spotting method that detects and recognizes text in a single shot. We demonstrate state of the art performance in most scene text retrieval benchmarks. Moreover, we show that our scene text retrieval method yields equally good results for in-dictionary and out-of-dictionary (never before seen) text queries. Finally, we show that the resulting method is significantly faster than any state of the art approach for word spotting in scene images.

The proposed architecture is based on YOLO\cite{redmon2016you,redmon2016yolo9000}, a well known single shot object detector which we recast as a PHOC (Pyramidal Histogram Of Characters) \cite{almazan2014word,sudholt2016phocnet} predictor, thus being able to effectively perform word detection and recognition at the same time. 
The main contribution of this paper is the demonstration that using PHOC as a word representation instead of a direct word classification over a closed dictionary, provides an elegant mechanism to generalize to any text string, allowing the method to tackle efficiently out-of-dictionary queries. By learning to predict PHOC representations of words the proposed model is able to transfer the knowledge acquired from training data to represent words it has never seen before.

The remainder of this paper is organized as follows. Section~\ref{sec:related} presents an overview of the state of the art in scene text understanding tasks, Section~\ref{sec:architecture} describes the proposed architecture for single shot scene text retrieval. Section~\ref{sec:results} reports the experiments and results obtained on different benchmarks for scene text based image retrieval. Finally, the conclusions and pointers to further research are given in Section~\ref{sec:conclusion}.

\section{Related work}
\label{sec:related}

The first attempts at recognizing text in scene images divided the problem in two distinguished steps, text detection and text recognition. For instance, in the work of Jaderberg~\etal~\cite{jaderberg2016reading} scene text segmentation was performed by a text proposals mechanism that was later refined by a CNN that regressed the correct position of bounding boxes. Afterwards, those bounding boxes were inputed to a CNN that classified them in terms of a predefined vocabulary. Gupta~\etal\cite{gupta2016synthetic} followed a similar strategy by first using a Fully Convolutional Regression Network for detection and the same classification network than Jaderberg for recognition. Liao~\etal\cite{liao2017textboxes,liao2018textboxes++} used a modified version of the SSD~\cite{liu2016ssd} object detection architecture adapted to text and then a CRNN~\cite{shi2017end} for text recognition. However, breaking the problem into two separate and independent steps presented an important drawback since detection errors might significantly hinder the further recognition step. Recently, end-to-end systems that approach the problem as a whole have gained the attention of the community. Since the segmentation and recognition tasks are highly correlated from an end to end perspective, in the sense that learned features can be used to solve both problems, researchers started to jointly train their models. Buvsta~\etal\cite{buvsta2017deep} proposed to use a Fully Convolutional Neural Network for text detection and another module that employed a CTC (Connectionist Temporal Classification) for text recognition. Both modules were first trained independently and further joined together in order to make an end-to-end trainable architecture. Li~\etal\cite{li2017towards} proposed a pipeline that included a CNN to obtain text region proposals followed by a region feature encoding module that is the input to an LSTM to detect text. The detected regions are the input to another LSTM which outputs features to be decoded by a LSTM with attention to recognize the words. In that sense, we strongly believe that single shot object detection paradigms such as YOLO~\cite{redmon2016yolo9000} can bring many benefits to the field of scene text recognition by having a unique architecture that is able to locate and recognize the desired text in an unique step.

However, the scene text retrieval problem slightly differs from classical scene text recognition applications. In a retrieval scenario the user should be able to cast whatever textual query he wants to retrieve, whereas most of recognition approaches are based on using a predefined vocabulary of the words one might find within scene images. For instance, both Mishra~\etal\cite{mishra2013image}, who introduced the scene text retrieval task, and Jaderberg~\etal\cite{jaderberg2016reading}, use a fixed vocabulary to create an inverted index which contains the presence of a word in the image. Such approach obviously limits the user that does not have the freedom to cast out of vocabulary queries. In order to tackle such problem, text string descriptors based on n-gram frequencies, like the PHOC descriptor, have been successfully used for word spotting applications~\cite{aldavert13,almazan2014word,ghosh2015query}. By using a vectorial codification of text strings, users can cast whatever query at processing time without being restricted to specific word sets.

\section{Single shot word spotting architecture}
\label{sec:architecture}
\begin{figure}
\centering
\resizebox{120mm}{!}{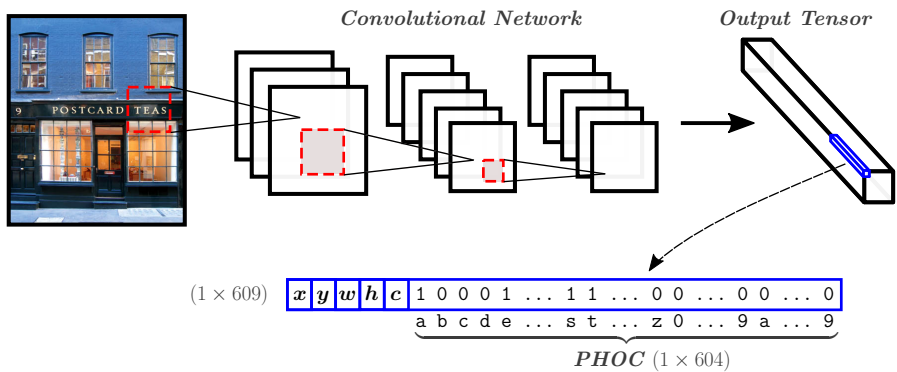}
\caption{Our Convolutional Neural Network predicts at the same time bounding box coordinates $x,y,w,h$, an objectness score $c$, and a pyramidal histogram of characters (PHOC) of the word in each bounding box.}
\label{fig:yolo-phoc}
\end{figure}

The proposed architecture, illustrated in Figure~\ref{fig:yolo-phoc}, consists in a single shot CNN model that predicts at the same time bounding boxes and a compact text representation of the words within them. To accomplish this we adapt the YOLOv2 object detection model \cite{redmon2016you,redmon2016yolo9000} and recast it as a PHOC~\cite{almazan2014word} predictor. Although the proposed method can be implemented on top of other object detection frameworks we opted for YOLOv2 because it can be up to $10\times$ faster than two-stage frameworks like Faster R-CNN~\cite{ren2015faster}, and processing time is critical for us since we aim at processing images at high resolution to correctly deal with small text.

The YOLOv2 architecture is composed of 21 convolutional layers with a leaky ReLU activation and batch normalization [7] and 5 max pooling layers. It uses $3 \times 3$ filters and double the number of channels after every pooling step as in VGG models [17], but also uses $1 \times 1$ filters interspersed between $3 \times 3$ convolutions to compress the feature maps as in [9]. The backbone includes a pass-through layer from the second convolution layer and is followed by a final $1 \times 1$ convolutional layer with a linear activation with the number of filters matching the desired output tensor size for object detection. For example, in the PASCAL VOC challenge dataset (20 object classes) it needs $125$ filters to predict $5$ boxes with $4$ coordinates each, 1 objectness value, and $20$ classes per box ($(4+1+20) \times 5 = 125$). The resulting model achieves state of the art in object detection, has a smaller number of parameters than other single shot models, and features real time object detection.

A straightforward application of the YOLOv2 architecture to the word spotting task would be to treat each possible word as an object class. This way the one hot classification vectors in the output tensor would encode the word class probability distribution among a predefined list of possible words (the dictionary) for each bounding box prediction. The downside of such an approach is that we are limited in the number of words the model can detect. For a dictionary of $20$ words the model would theoretically perform as well as for the $20$ object classes of the PASCAL dataset, but training for a larger dictionary (e.g. the list of $100,000$ most frequent words from the English vocabulary~\cite{jaderberg2016reading}) would require a final layer with $500,000$ filters, and a tremendous amount of training data if we want to have enough samples for each of the $100,000$ classes. Even if we could manage to train such a model, it would still be limited to the dictionary size and not able to detect any word not present on it.

\begin{figure}[ht]
\centering
\includegraphics[scale=.40]{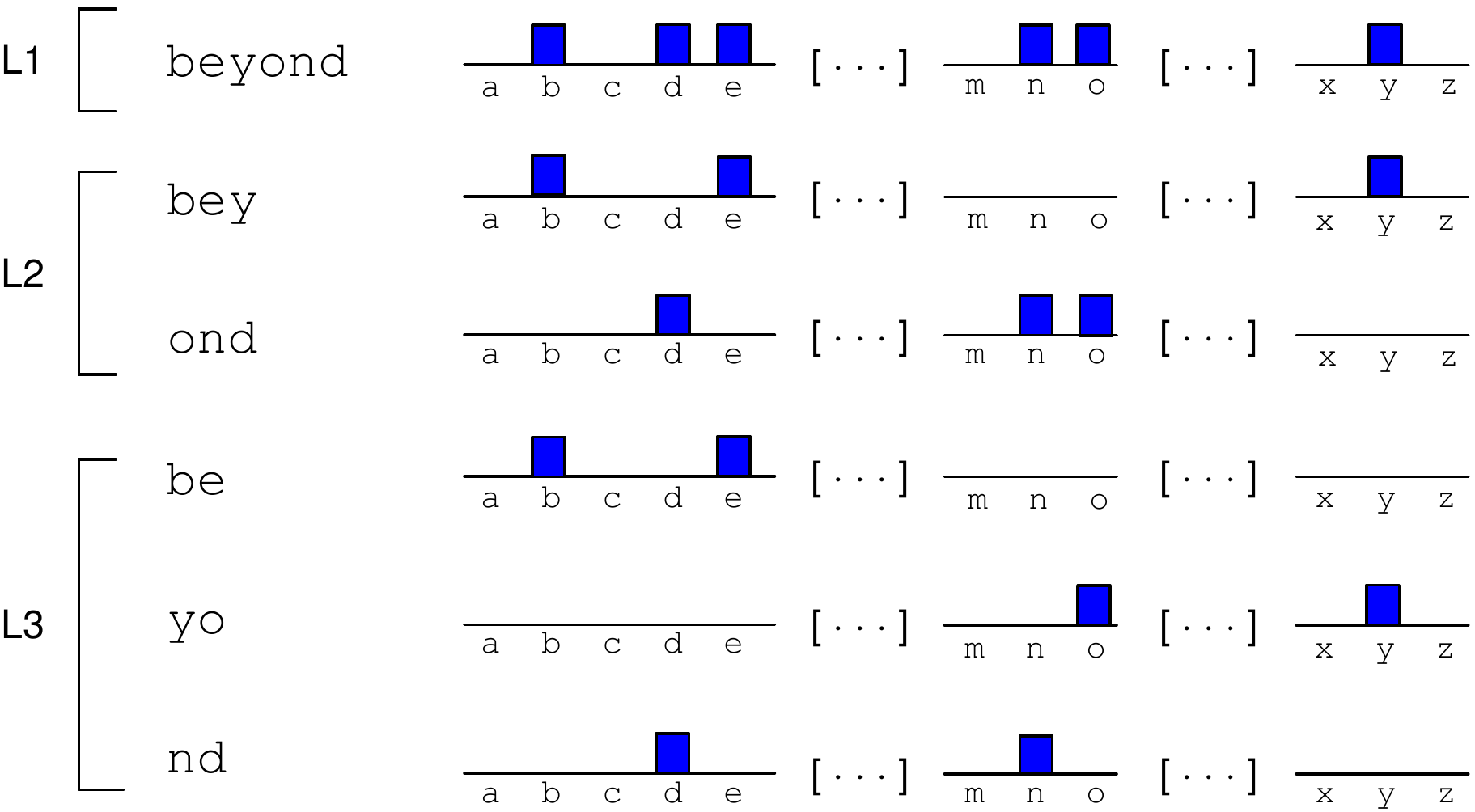}
\caption{Pyramidal histogram of characters (PHOC)~\cite{almazan2014word} of the word ``beyond'' at levels 1, 2, and 3. The final PHOC representation is the concatenation of these partial histograms.}
\label{fig:phoc}
\end{figure}

Instead of the fixed vocabulary approach we would like to have a model that is able to generalize to words that were not seen at training time. This is the rationale behind casting the network as a PHOC predictor. PHOC~\cite{almazan2014word} is a compact representation of text strings that encodes if a specific character appears in a particular spatial region of the string (see Figure \ref{fig:phoc}). Intuitively a model that effectively learns to predict PHOC representations will implicitly learn to identify the presence of a particular character in a particular region of the bounding box by learning character attributes independently. This way the knowledge acquired from training data can be transfered at test time for words never observed during training, because the presence of a character at a particular location of the word translates to the same information in the PHOC representation independently of the other characters in the word. Moreover, the PHOC representation offers unlimited expressiveness (it can represent any word) with a fixed length low dimensional binary vector ($604$ dimensions in the version we use).

In order to adapt the YOLOv2 network for PHOC prediction we need to address some particularities of this descriptor. First, since the PHOC representation is not a one hot vector we need to get rid of the softmax function used by YOLOv2 in the classification output. Second, since the PHOC is a binary representation it makes sense to squash the network output corresponding to the PHOC vector to the range {0...1}. To accomplish this, a sigmoid activation function was used in the last layer. Third, we propose to modify the original YOLOv2 loss function in order to help the model through the learning process. The original YOLOv2 model optimizes the following multi-part loss function:

\begin{equation}
L (b,C,c,\hat{b},\hat{C},\hat{c}) = \lambda_{box} L_{box} (b, \hat{b}) + L_{obj} (C, \hat{C}, \lambda_{obj}, \lambda_{noobj}) + \lambda_{cls} L_{cls} (c, \hat{c}) 
\end{equation}

\noindent
where $b$ is a vector with coordinates' offsets to an anchor bounding box, $C$ is the probability of that bounding box containing an object, $c$ is the one hot classification vector, and the three terms $L_{box}$, $L_{obj}$, and $L_{cls}$ are respectively independent losses for bounding box regression, objectness estimation, and  classification. All the aforementioned losses are essentially the sum-squared errors of ground truth ($b,C,c$) and predicted ($\hat{b},\hat{C},\hat{c}$) values. In the case of PHOC prediction, with $c$ and $\hat{c}$ being binary vectors but with an unrestricted number of $1$ values we opt for using a cross-entropy loss function in $L_{cls}$ as in a multi-label classification task:

\begin{equation}
L_{cls}(c, \hat{c}) =  \frac{-1}{N} \sum_{n=1}^N \left[ c_n \log(\hat{c}_n) + (1 - c_n) \log(1 - \hat{c}_n) \right]
\end{equation}
\noindent
where $N$ is the dimensionality of the PHOC descriptor.

Similarly as in~\cite{redmon2016you} the combination of the sum-squared errors $L_{box}$ and $L_{obj}$ with the cross-entropy loss $L_{cls}$ is controlled by the scaling parameters $\lambda_{box}$, $\lambda_{obj}$, $\lambda_{noobj}$, and $\lambda_{cls}$.

\begin{figure}
\centering
\begin{tabular}{c c c c c}
\includegraphics[width=.3\textwidth]{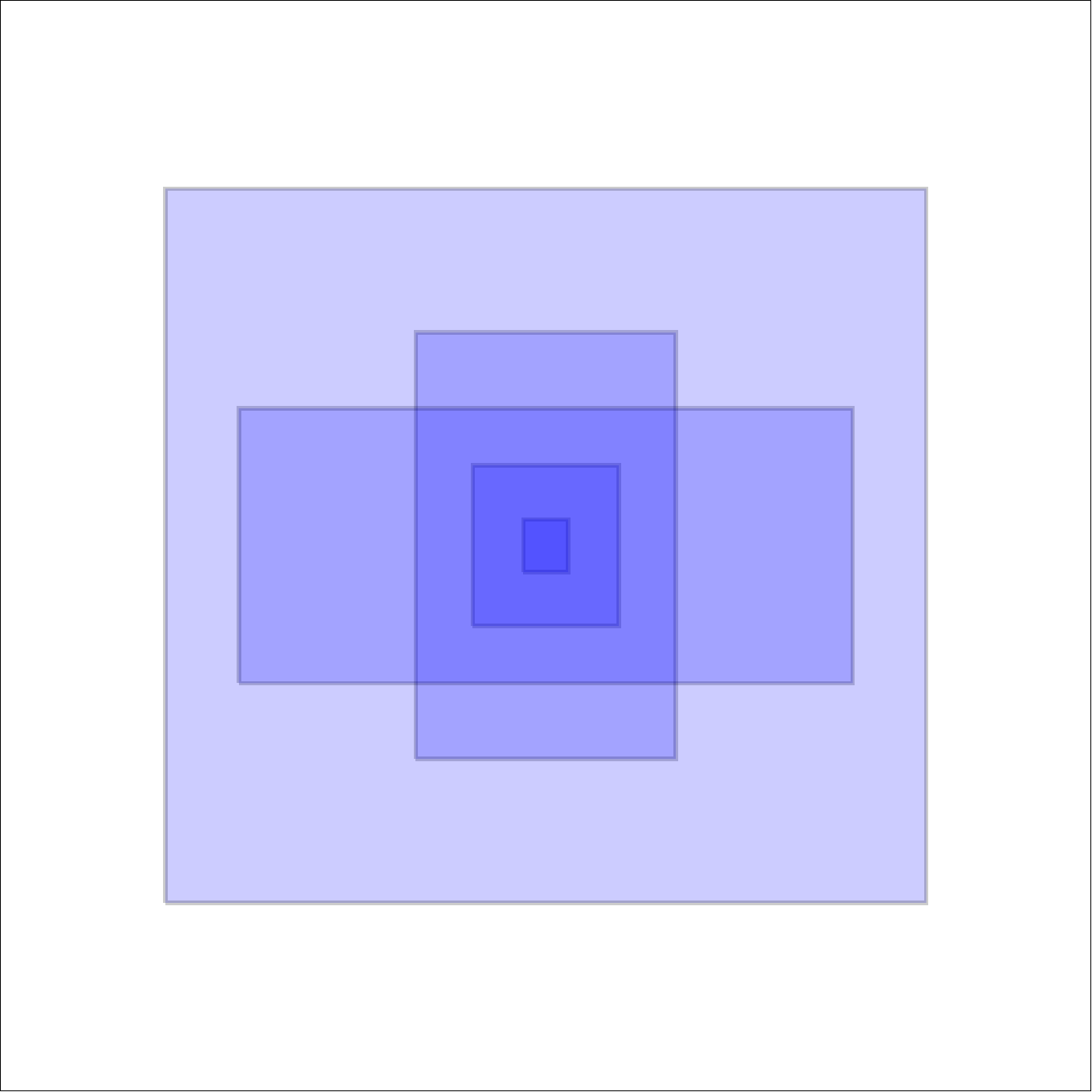} & \hspace{0.3cm} & \includegraphics[width=.3\textwidth]{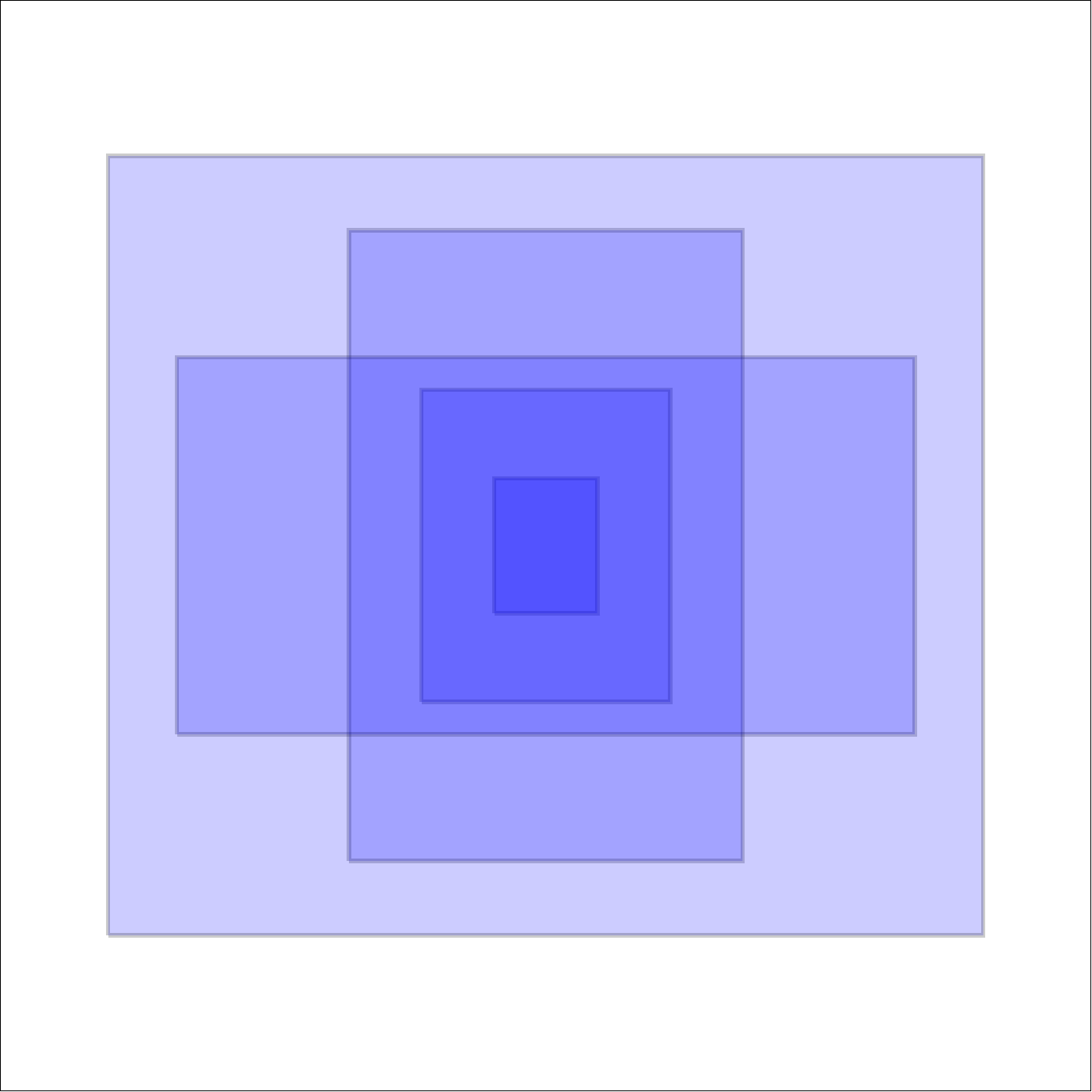} & \hspace{0.3cm} & \includegraphics[width=.3\textwidth]{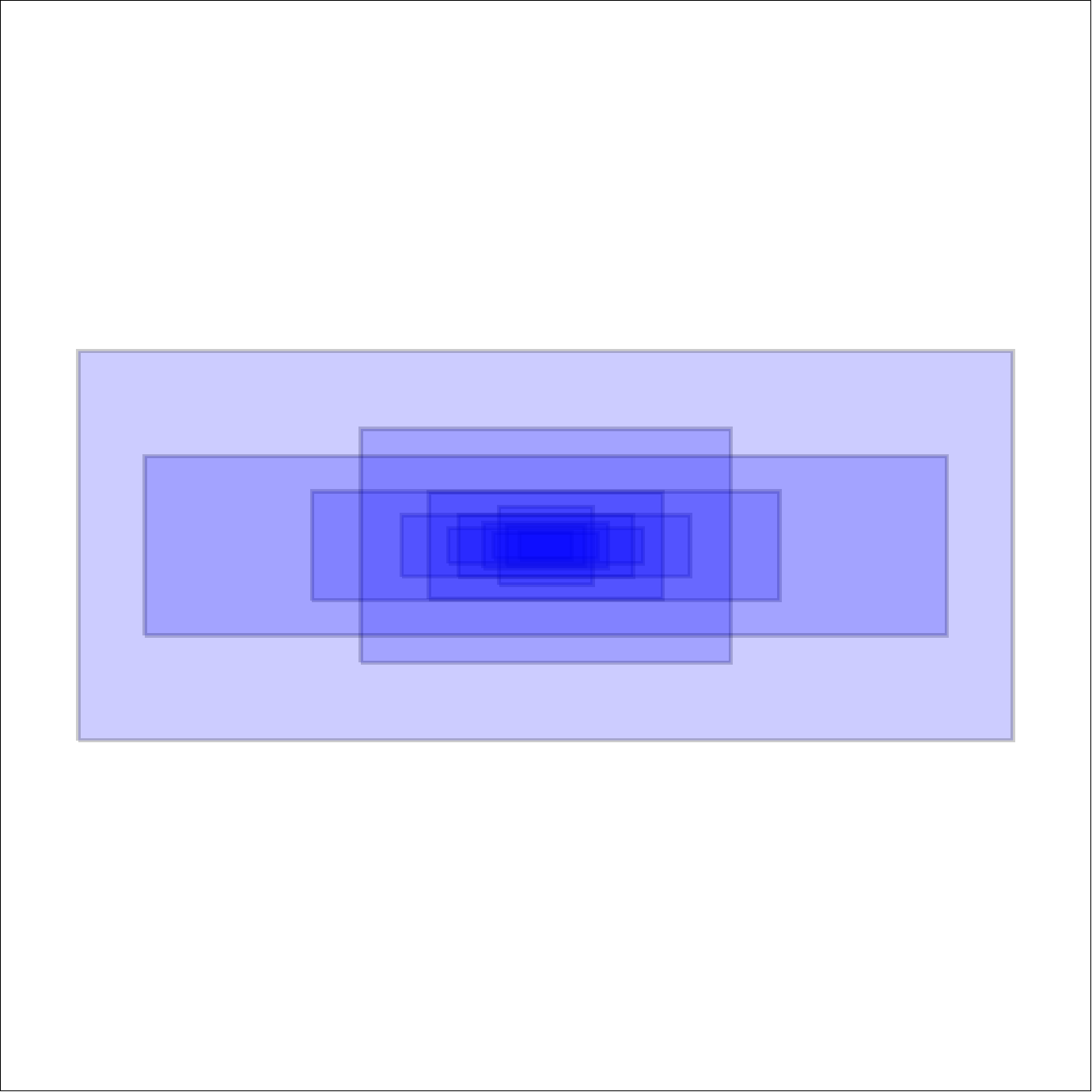} \\
a) & & b) & & c)\\
\end{tabular}
\caption{Anchor boxes used in the original YOLOv2 model for object detection in COCO (a) and PASCAL (b) datasets. (c) Our set of anchor boxes for text detection.}
\label{fig:anchors}
\end{figure}

Apart of the modifications made so far on top of the original YOLOv2 architecture we also changed the number, the scales, and the aspect ratios of the pre-defined anchor boxes used by the network to predict bounding boxes. Similarly as in~\cite{redmon2016you} we have found the ideal set of anchor boxes $B$ for our training dataset by requiring that for each bounding box annotation there exists at least one anchor box in $B$ with an intersection over union of at least $0.6$. Figure~\ref{fig:anchors} illustrates the $13$ bounding boxes found to be better suited for our training data and their difference with the ones used in object detection models.

At test time, our model provides a total of $W/32 \times H/32 \times 13$ bounding box proposals, with $W$ and $H$ being the image input size, each one of them with an objectness score ($\hat{C}$) and a PHOC prediction ($\hat{c}$). The original YOLOv2 model filters the bounding box candidates with a detection threshold $\tau$ considering that a bounding box is a valid detection if $\hat{C} max(\hat{c}) \geq \tau$. If the threshold condition is met, a non-maximal suppression (NMS) strategy is applied in order to get rid of overlapping detections of the same object. In our case the threshold is applied only on the objectness score ($\hat{C}$) but with a much smaller value ($\tau = 0.0025$) than in the original model ($\tau \approx 0.2$), and we do not apply NMS. The reason is that any evidence of the presence of a word, even if it is small, it may be beneficial in terms of retrieval if its PHOC representation has a small distance to the PHOC of the queried word. With this threshold we generate an average of $60$ descriptors for every image in the dataset and all of them conform our retrieval database.

In this way, the scene text retrieval of a given query word is performed with a simple nearest neighbor search of the query PHOC representation over the outputs of the CNN in the entire image database. While the distance between PHOCs is usually computed using the cosine similarity, we did not find any noticeable downside on using an Euclidean distance for the nearest neighbor search. 

\subsection{Implementation details}

\begin{figure*}[t]
\centering
\includegraphics[width=1\linewidth]{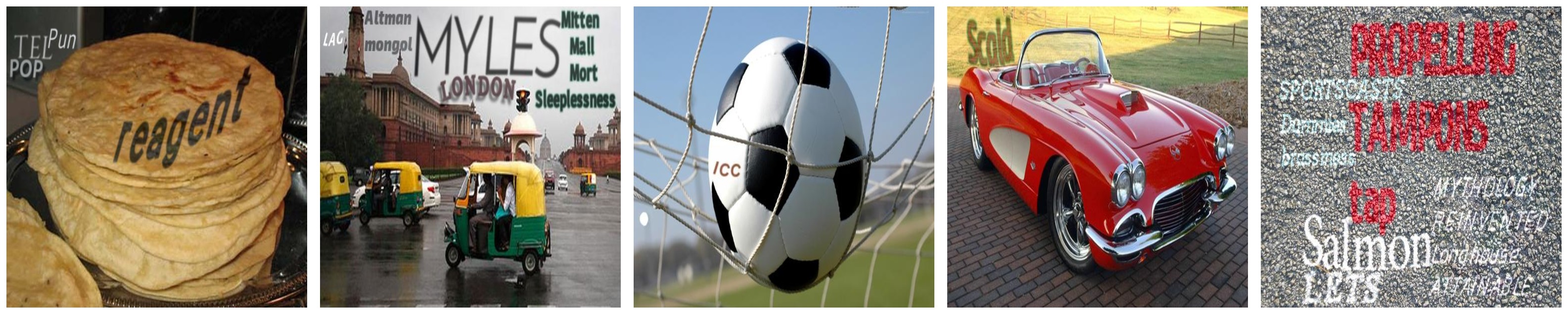}
\caption{Synthetic training data generated with a modified version of the method of Gupta~\etal\cite{gupta2016synthetic}. We make use of a custom dictionary with the 90K most frequent English words, and restrict the range of random rotation to 15 degrees.}
\label{fig:synth_data}
\end{figure*}

We have trained our model in a modified version of the synthetic dataset of Gupta~\etal\cite{gupta2016synthetic}. First the dataset generator has been evenly modified to use a custom dictionary with the 90K most frequent English words, as proposed by Jaderberg \etal\cite{jaderberg2016reading}, instead of the Newsgroup20 dataset~\cite{newsgroup20} dictionary originally used by Gupta~\etal The rationale was that in the original dataset there was no control about word occurrences, and the distribution of word instances had a large bias towards stop-words found in newsgroups' emails. Moreover, the text corpus of the Newsgroup20 dataset contains words with special characters and non ASCII strings that we do not contemplate in our PHOC representations. Finally, since the PHOC representation of a word with a strong rotation does not make sense under the pyramidal scheme employed, the dataset generator was modified to allow rotated text up to 15 degrees. This way we generated a dataset of 1 million images for training purposes. Figure \ref{fig:synth_data} shows a set of samples of our training data.

The model was trained for 30 epochs of the dataset using SGD with a batch size of $64$, an initial learning rate of $0.001$, a momentum of $0.9$, and a decay of $0.0005$. We initialize the weights of our model with the YOLOv2 backbone pre-trained on Imagenet. During the firsts 10 epochs we train the model only for word detection, without backpropagating the loss of the PHOC prediction and using a fixed input size of $448 \times 448$. On the following 10 epochs we start learning the PHOC prediction output with the $\lambda_{cls}$ parameter set to $1.0$. After that, we continue learning for 10 more epochs with a learning rate of $0.0001$ and setting the parameters $\lambda_{box}$ and $\lambda_{cls}$ to $5.0$ and $0.015$ respectively. At his point we also adopted a multi-resolution training, by randomly resizing the input images among $14$ possible sizes in the range from $352 \times 352$ to $800 \times 800$, and we added new samples in our training data. In particular, the added samples were the $1,233$ training images of the ICDAR2013~\cite{karatzas2013icdar} and ICDAR2015~\cite{karatzas2015icdar} datasets.  During the whole training process we used the same basic data augmentation as in~\cite{redmon2016you}.

\section{Experiments and results}
\label{sec:results}
In this section we present the experiments and results obtained on different standard benchmarks for text based image retrieval. First we describe the datasets used throughout our experiments, after that we present our results and compare them with the published state-of-the-art. Finally we discuss the scalability of the proposed retrieval method.

\subsection{Datasets}

\subsubsection{The IIIT Scene Text Retrieval (STR)} \cite{mishra2013image} dataset is a scene text image retrieval dataset composed of $10,000$ images collected from the Google image search engine and Flickr. The dataset has $50$ predefined query words and for each of them a list of $10-50$ relevant images (that contain the query word) is provided. It is a challenging dataset where relevant text appears in many different fonts and styles, and from different view points, among many distractors (images without any text).

\subsubsection{The IIIT Sports-10k dataset} \cite{mishra2013image} is another scene text retrieval dataset composed of $10,000$ images extracted from sports video clips. It has $10$ predefined query words with their corresponding relevant images' lists. Scene text retrieval in this dataset is specially challenging because images are low resolution and often noisy or blurred, with small text generally located on advertisements signboards.

\subsubsection{The Street View Text (SVT) dataset} \cite{wang2011end} is comprised of images harvested from Google Street View where text from business signs and names appear. It contains more than $900$ words annotated in $350$ different images. In our experiments we use the official partition that splits the images in a train set of $100$ images and a test set of $249$ images. This dataset also provides a lexicon of 50 words per image for recognition purposes, but we do not make use of it. For the image retrieval task we consider as queries the $427$ unique words annotated on the test set. 

\subsection{Scene text retrieval}

In the scene text retrieval task, the goal is to retrieve all images that contain instances of the query words in a dataset partition.
Given a query, the database elements are sorted with respect to the probability of containing the queried word. We use the mean average precision as the accuracy measure, which is the standard measure of performance for retrieval tasks and is essentially equivalent to the area below the precision-recall curve. Notice that, since the system always returns a ranked list with all the images in the dataset, the recall is always $100\%$. An alternative performance measure consist in considering only the top-$n$ ranked images and calculating the precision at this specific cut-off point ($P@n$).  

Table~\ref{tab:results_retrieval} compares the proposed method to previous state of the art for text based image retrieval on the IIIT-STR, Sports-10K, and SVT datasets. We show the mean average precision (mAP) and processing speed for the same trained model using two different input sizes ($576 \times 576$ and $608 \times 608$), and a multi-resolution version that combines the outputs of the model at three resolutions ($544$, $576$ and $608$). Processing time has been calculated using a Titan X (Pascal) GPU with a batch size of $1$. We appreciate that our method outperforms previously published methods in two of the benchmarks while it shows a competitive performance on the SVT dataset. In order to compare with state-of-the-art end-to-end text recognition methods, we also provide a comparison with pre-trained released versions of the models of Bu\v{s}ta \etal\cite{buvsta2017deep} and He \etal\cite{he2018single}. For recognition-based results the look-up is performed by a direct matching between the query and the text detected by each model. Even when making use of a predefined word dictionary to filter results, our method, which is dictionary-free, yields superior results. 
Last, we compared against a variant of  
He \etal\cite{he2018single} but this time both queries and the model's results are first transformed to PHOC descriptors and the look-up is based on similarity on PHOC space. It can be seen that the PHOC space does not offer any advantage to end-to-end recognition methods. 

\begin{table}[h]
\caption{Comparison to previous state of the art for text based image retrieval: mean average precision (mAP) for IIIT-STR, and Sports-10K, and SVT datasets. (*) Results reported by Mishra et al. in \cite{mishra2013image}, not by the original authors. ($\dagger$) Results computed with publicly available code from the original authors.}
\label{tab:results_retrieval}
\centering
\begin{tabularx}{\textwidth}{ l Y Y Y Y}
\toprule
Method &  STR (mAP) & Sports (mAP) & SVT (mAP) & fps\\
\midrule
SWT \cite{epshtein2010detecting}+ Mishra et al. \cite{mishra2012top} & - & - & 19.25 \\
Wang \etal \cite{wang2011end} & - & - & 21.25* \\
TextSpotter \cite{neumann2012real} & - & - & 23.32* & 1.0\\
Mishra \etal \cite{mishra2013image}& 42.7 & - & 56.24 & 0.1 \\
Ghosh \etal \cite{ghosh2015efficient} & - & - & 60.91 \\
Mishra \cite{mishra2016understanding} & 44.5 & - & 62.15 & 0.1 \\
Almaz{\'a}n  \etal \cite{almazan2014word} & - & - & 79.65 & \\
TextProposals \cite{gomez2017textproposals} + DictNet  \cite{jaderberg2014synthetic} & 64.9$^\dagger$ & 67.5$^\dagger$ & 85.90$^\dagger$ & 0.4\\
Jaderberg \etal \cite{jaderberg2016reading} & 66.5 & 66.1 & \bf{86.30} & 0.3\\
Bu\v{s}ta \etal \cite{buvsta2017deep} & 62.94$^\dagger$& 59.62$^\dagger$& 69.37$^\dagger$& 44.21 \\
He \etal \cite{he2018single} & 50.16$^\dagger$& 50.74$^\dagger$& 72.82$^\dagger$& 1.25 \\
He \etal \cite{he2018single} (with dictionary) & 66.95$^\dagger$& 74.27$^\dagger$& 80.54$^\dagger$& 2.35 \\
He \etal \cite{he2018single} (PHOC) & 46.34$^\dagger$& 52.04$^\dagger$& 57.61$^\dagger$& 2.35 \\
\midrule
Proposed ($576 \times 576$) & \bf{68.13} & 72.99 & 82.02 & \bf{53.0}\\ 
Proposed ($608 \times 608$) & \bf{69.83} & 73.75 & 83.74 & 43.5\\
Proposed (multi-res.) & \bf{71.37} & \bf{74.67} & 85.18 & 16.1\\ 
\bottomrule
\end{tabularx}
\end{table}


Table~\ref{tab:results_retrieval2} further compares the proposed method to previous state of the art by the precisions at 10 (P@10) and 20 (P@20) on the Sports-10K dataset.

\begin{table}[h!]
\caption{Comparison to previous state of the art for text based image retrieval: precision at n (P@n) for Sports-10K dataset.}
\label{tab:results_retrieval2}
\centering
\begin{tabularx}{\textwidth}{ X Y Y }
\toprule
 Method & Sports-10K (P@10) & Sports-10K (P@20)\\
\midrule
Mishra \etal \cite{mishra2013image}& 44.82 & 43.42 \\
Mishra \cite{mishra2016understanding} & 47.20 & 46.25 \\
Jaderberg \etal \cite{jaderberg2016reading} & 91.00 & \bf{92.50}\\
\midrule
Proposed  ($576 \times 576$) & 91.00 & 90.50\\ 
Proposed (multi-res.) & \bf{92.00} & 90.00 \\ 
\bottomrule
\end{tabularx}
\end{table}

\begin{figure*}
\centering
\includegraphics[width=1\linewidth]{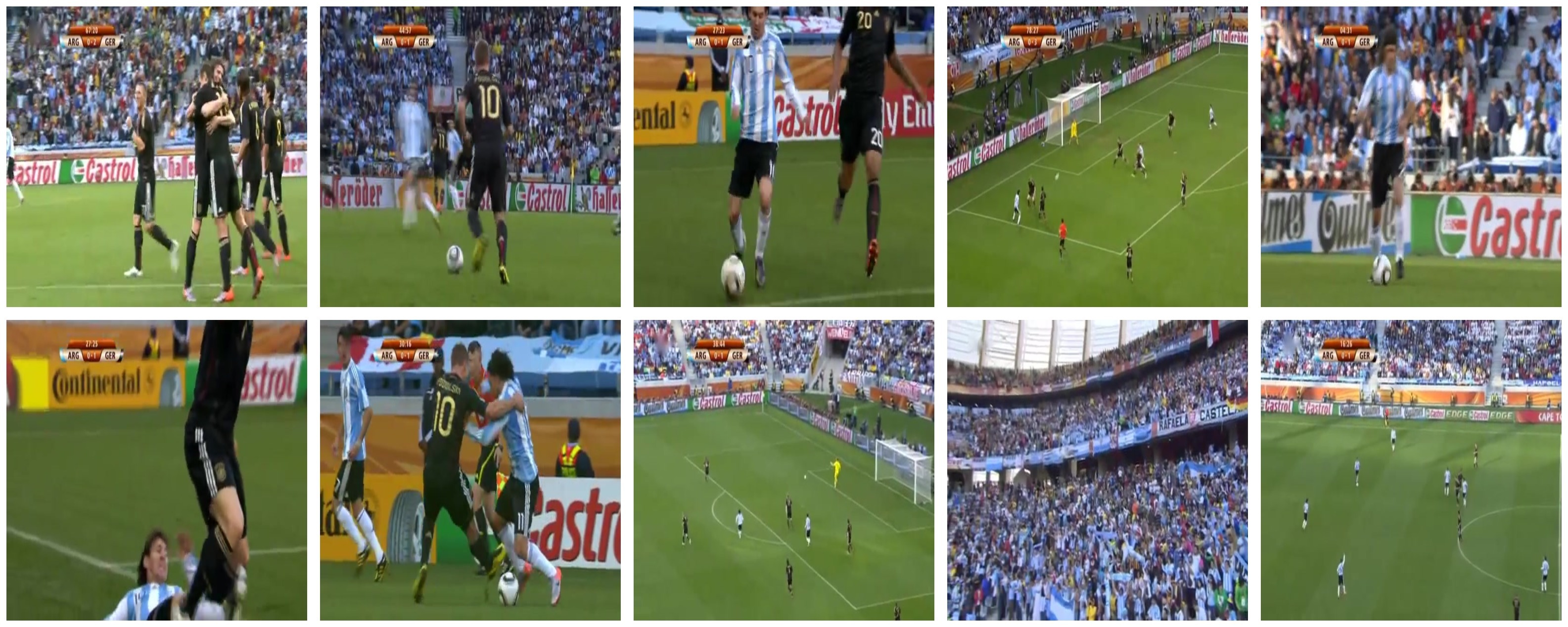}
\caption{Top 10 ranked images for the query ``castrol''. Our model has not seen this word at training time.}
\label{fig:castrol}
\end{figure*}

In Table~\ref{tab:results_retrieval_per_query} we show per-query mean average precision and precisions at $10$ and $20$ for the Sports-10K dataset. The low performance for the query ``castrol'' in comparison with the rest may initially be attributed to the fact that it is the only query word not seen by our model at training time. However, by visualizing the top-10 ranked images for this query, shown in Figure~\ref{fig:castrol} we can see that the dataset has many unannotated instances of ``castrol''. The real P@10 of our model is in fact $90\%$ and not $50\%$. It appears that the annotators did not consider occluded words, while our model is able to retrieve images with partial occlusions in a consistent manner. Actually, the only retrieved image among the top-10 without the ``castrol'' word contains an instance of ``castel''.  By manual inspection we have computed P@10 and P@20 to be $95.0$ and $93.5$ respectively.

\begin{table}[h]
\caption{Sports-10K per-query average precision (AP), P@10, and P@20 scores.}
\label{tab:results_retrieval_per_query}
\centering
\begin{tabularx}{\textwidth}{ X | Y | Y | Y | Y | Y | Y | Y | Y | Y | Y }
\toprule
& \rotatebox[origin=c]{90}{\parbox[c]{1.1cm}{\centering adidas}} & \rotatebox[origin=c]{90}{\parbox[c]{1.1cm}{\centering castrol}} & \rotatebox[origin=c]{90}{\parbox[c]{1.1cm}{\centering duty}} & \rotatebox[origin=c]{90}{\parbox[c]{1.1cm}{\centering free}} & \rotatebox[origin=c]{90}{\parbox[c]{1.1cm}{\centering hyundai}} & \rotatebox[origin=c]{90}{\parbox[c]{1.1cm}{\centering nokia}} & \rotatebox[origin=c]{90}{\parbox[c]{1.1cm}{\centering pakistan}} & \rotatebox[origin=c]{90}{\parbox[c]{1.1cm}{\centering pepsi}} & \rotatebox[origin=c]{90}{\parbox[c]{1.1cm}{\centering reliance}} & \rotatebox[origin=c]{90}{\parbox[c]{1.1cm}{\centering sony}} \\
\midrule
AP & 94 & 16 & 74 & 61 & 77 & 75 & 92 & 70 & 89 & 89 \\
P@10 & 100 & 50 & 100 & 90 & 100 & 80 & 100 & 90 & 100 & 90\\ 
P@20 & 100 & 55 & 100 & 85 & 100 & 85 & 100 & 95 & 100 & 90\\
\bottomrule
\end{tabularx}
\end{table}

\begin{figure*}
\centering
\includegraphics[width=1\linewidth]{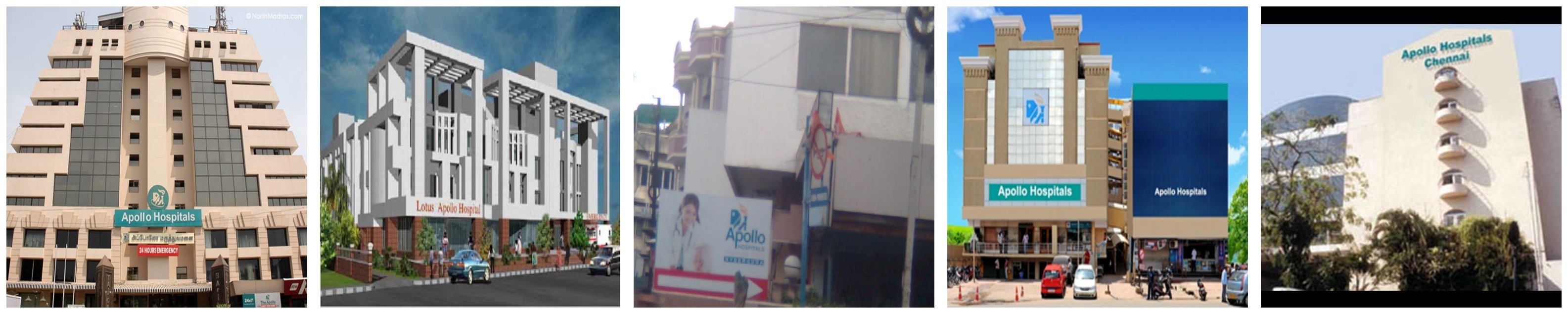}\\
\includegraphics[width=1\linewidth]{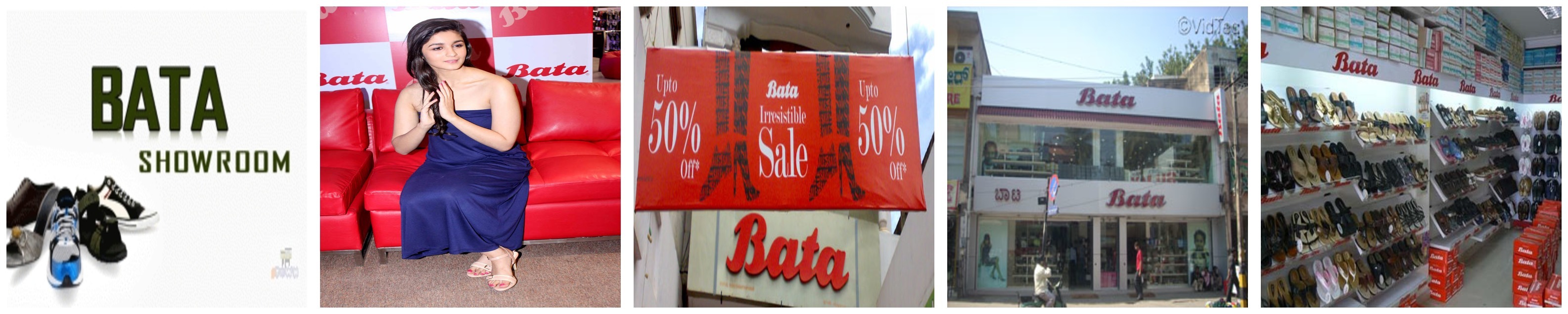}\\
\includegraphics[width=1\linewidth]{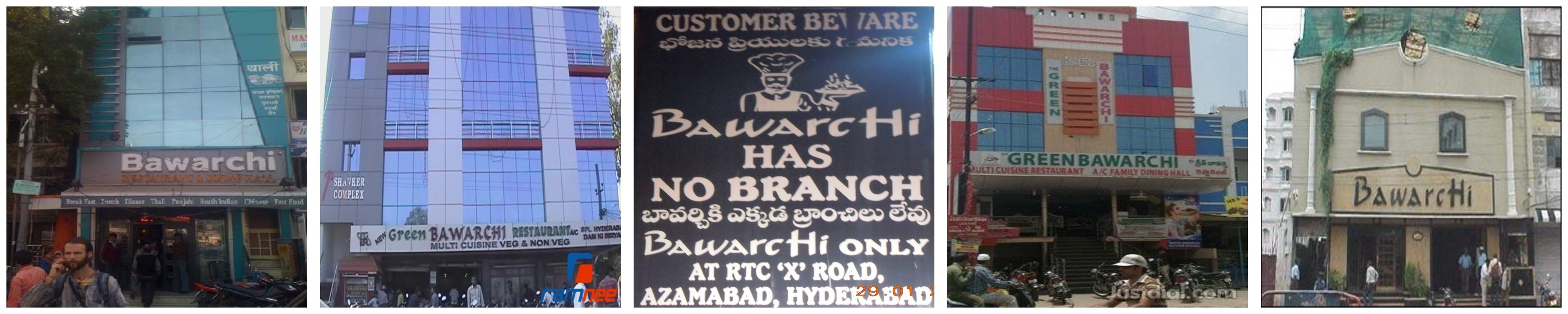}\\
\includegraphics[width=1\linewidth]{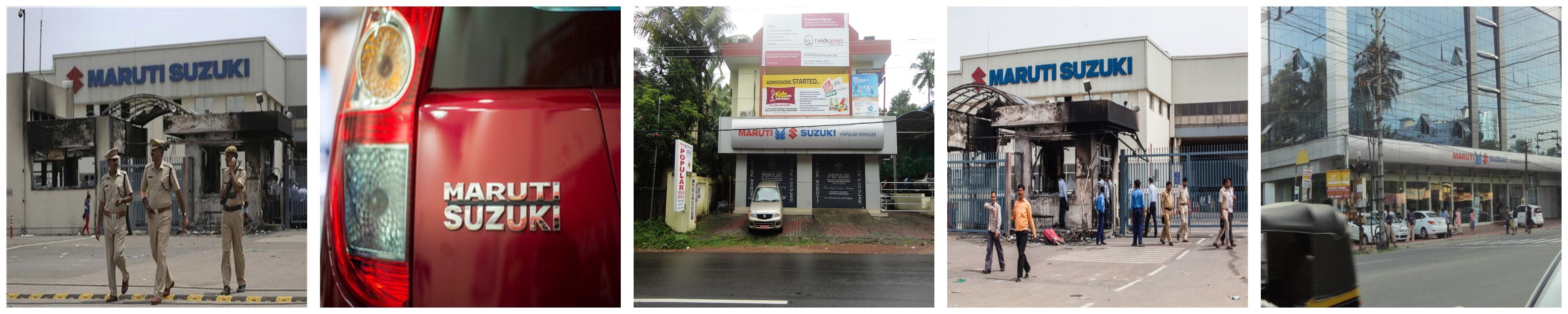}\\
\includegraphics[width=1\linewidth]{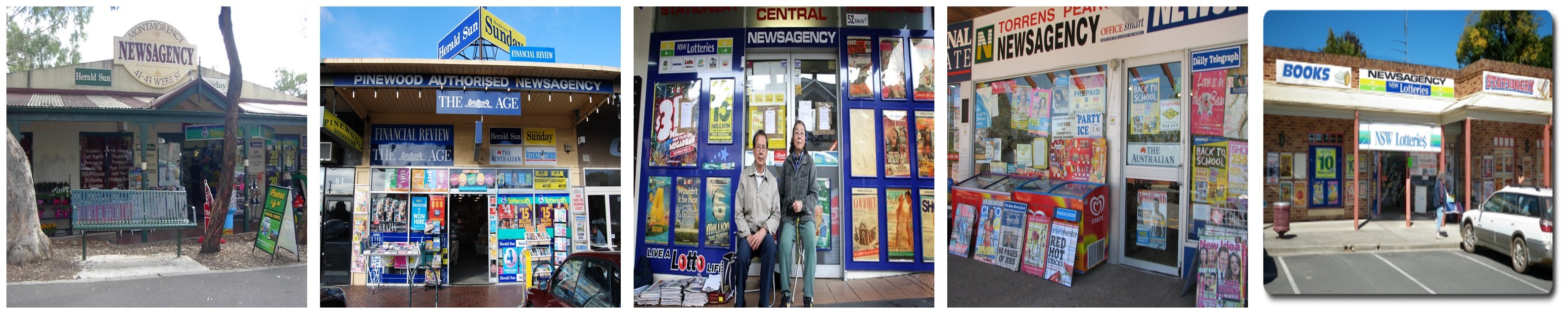}\\
\includegraphics[width=1\linewidth]{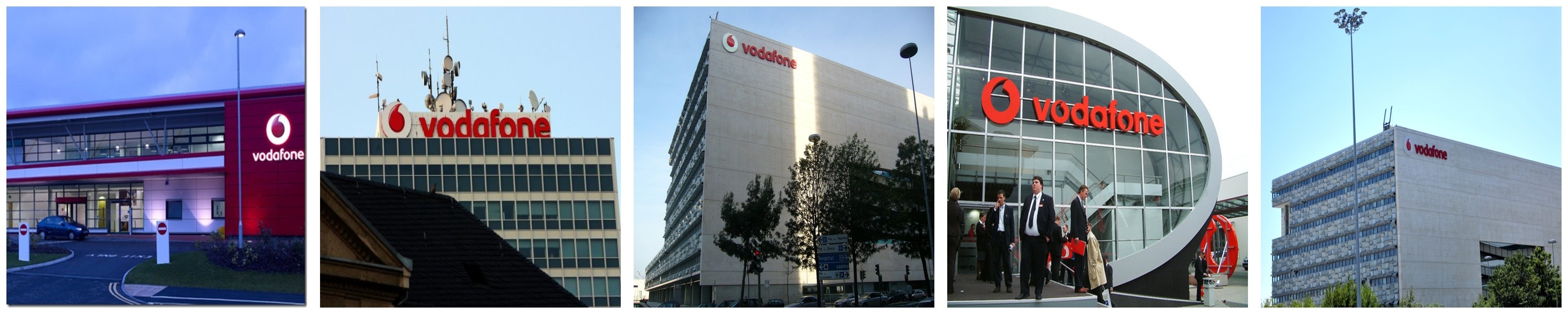}
\caption{From top to bottom, top-5 ranked images for the queries ``apollo'', ``bata'', ``bawarchi'', ``maruti'', ``newsagency'', and ``vodafone''. Although our model has not seen this words at training time it is able to achieve a $100\%$ P@5 for all of them.}
\label{fig:str_oov}
\end{figure*}

Overall, the performance exhibited with the ``castrol'' query is a very important result, since it demonstrates that our model is able to generalize the PHOC prediction for words that has never seen at training time, and even to correctly retrieve them under partial occlusions. We found further support for this claim by analyzing the results for the six IIIT-STR query words that our model has not seen during training. Figure~\ref{fig:str_oov} shows the top-5 ranked images for the queries ``apollo'', ``bata'', ``bawarchi'', ``maruti'', ``newsagency'', and ``vodafone''. In all of them  our model reaches a $100\%$ precision at $5$. In terms of mAP the results for these queries do not show a particular decrease when compared to those obtained with other words that are part of the training set, in fact in some cases they are even better. The mean average precision for the six words in question is $74.92$, while for the remaining $44$ queries is $69.14$. To further analyze our model's ability for recognizing words it has never seen at training time, we have done an additional experiment within a multi-lingual setup. For this we manually added some images with text in different Latin script languages (French, Italian, Catalan, and Spanish) to the IIIT-STR dataset. We have observed that our model, while being trained only using English words, was always able to correctly retrieve the queried text in any of those languages.

\begin{figure*}
\centering
\includegraphics[width=1\linewidth]{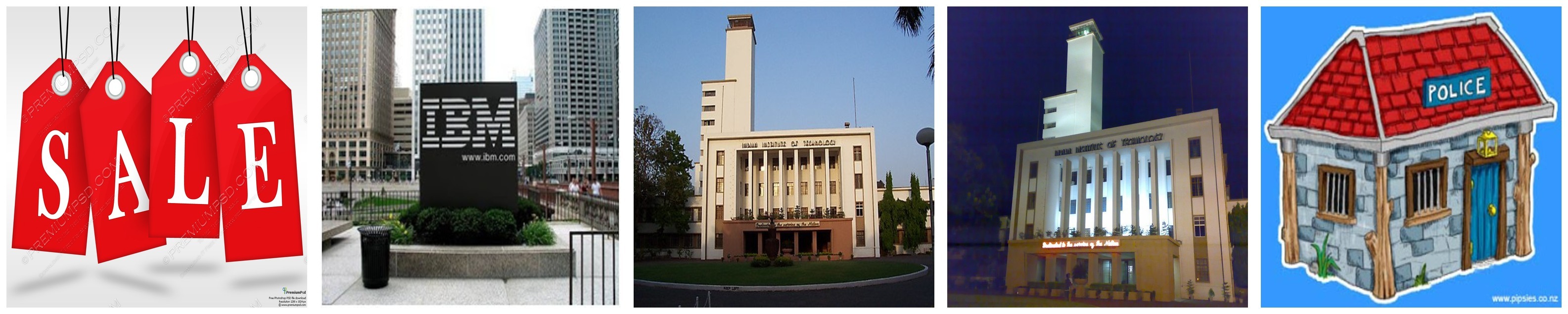}
\caption{Error analysis: most of the errors made by our model come from text instances with a particular style, font type, size, etc. that is not well represented in our training data.}
\label{fig:errors}
\end{figure*}

In order to analyze the errors made by our model we have manually inspected the output of our model as well as the ground truth for the five queries with a lower mAP on the IIIT-STR dataset: ``ibm'', ``indian'', ``institute'', ``sale'', and ``technology''. In most of these queries the low accuracy of our model can be explained in terms of having only very small and blurred instances in the database. In the case of ``ibm'', the characteristic font type in all instances of this word tends to be ignored by our model, and the same happens for some computer generated images (\ie non scene images) that contain the word ``sale''. Figure~\ref{fig:errors} shows some examples of those instances. All in all, the analysis indicates that while our model is able to generalize well for text strings not seen at training time it does not perform properly with text styles, fonts, sizes not seen before. Our intuition is that this problem can be easily alleviated with a richer training dataset.

\subsection{Retrieval speed analysis}
\label{sec:scalability}
To analyze the retrieval speed of the proposed system, we have run the retrieval experiments for the IIIT-STR and Sports-10K datasets with different approximate nearest neighbor (ANN) algorithms in a standard PC with an i7 CPU and 32Gb of RAM. In Table~\ref{tab:ann} we appreciate that those ANN methods, with a search time sublinear in the number of indexed samples, reach retrieval speeds a couple of orders of magnitude faster than the exact nearest neighbor search based on ball-trees without incurring in any significant loss of retrieval accuracy.

\begin{table}
\caption{Mean Average Precision and retrieval time performance (in seconds) of different approximate nearest neighbor algorithms on the IIIT-STR and Sports datasets.}
\label{tab:ann}
\centering
\begin{tabularx}{\textwidth}{ X c c c c c c }
\toprule
  & \multicolumn{3}{c}{IIIT-STR} & \multicolumn{3}{c}{Sports-10K}\\
\midrule
Algorithm & ~ mAP ~ & ~ secs ~ & \#PHOCs & ~ mAP ~ & ~ secs ~ & \#PHOCs \\
\midrule
Baseline (Ball tree) & 0.6983 & 0.4321 & 620K & 0.7375 & 0.6826 & 1M \\
Annoy (approx NN)~\cite{annoy} & 0.6883 & 0.0027 & 620K & 0.7284 & 0.0372 & 1M \\
HNSW (approx NN)~\cite{malkov2016efficient} & 0.6922 & 
0.0018 & 620K & 0.7247 & 0.0223 & 1M \\
Falconn LSH (approx NN)~\cite{andoni2015practical} & 0.6903 & 0.0151 & 620K & 0.7201 & 0.0178 & 1M \\
\bottomrule
\end{tabularx}
\end{table}



\section{Conclusion}
\label{sec:conclusion}
In this paper we detailed a real-time word spotting method, based on a simple architecture that allows it to detect and recognise text in a single shot and real-time speeds.

The proposed method significantly improves state of the art results on scene text retrieval on the IIIT-STR and Sports-10K datasets, while yielding comparable results to state of the art in the SVT dataset. Moreover, it can do so achieving faster speed compared to other state of the art methods.

Importantly, 
the proposed method is fully capable to deal with out-of-dictionary (never before seen) text queries, seeing its performance unaffected compared to query words previously seen in the training set.

This is due to the use of PHOC as a word representation instead of aiming for a direct word classification. It can be seen that the network is able to learn how to extract such representations efficiently, generalizing well to unseen text strings. Synthesizing training data with different characteristics could boost performance, and is one of the directions we will be exploring in the future along with investigating the use of word embeddings other than PHOC.

The code, pre-trained models, and data used in this work are made publicly available at \url{https://github.com/lluisgomez/single-shot-str}.

\section*{Acknowledgement}
This work has been partially supported by the Spanish research project TIN2014-52072-P, the CERCA Programme / Generalitat de Catalunya, the H2020 Marie Skłodowska-Curie actions of the European Union, grant agreement No 712949 (TECNIOspring PLUS), the Agency for Business Competitiveness of the Government of Catalonia (ACCIO), CEFIPRA Project 5302-1 and the project ``aBSINTHE - AYUDAS FUNDACI{\'O}N BBVA A EQUIPOS DE INVESTIGACION CIENTIFICA 2017. We gratefully acknowledge the support of the NVIDIA Corporation with the donation of the Titan X Pascal GPU used for this research.

\bibliographystyle{splncs}
\bibliography{eccv2018bib}

\begin{thebibliography}{10}

\bibitem{lin2014microsoft}
Lin, T.Y., Maire, M., Belongie, S., Hays, J., Perona, P., Ramanan, D.,
  Doll{\'a}r, P., Zitnick, C.L.:
\newblock Microsoft {COCO}: Common objects in context.
\newblock In: Proc. of the European Conference on Computer Vision, Springer
  (2014)  740--755

\bibitem{veit2016coco}
Veit, A., Matera, T., Neumann, L., Matas, J., Belongie, S.:
\newblock {COCO}-text: Dataset and benchmark for text detection and recognition
  in natural images.
\newblock arXiv preprint arXiv:1601.07140 (2016)

\bibitem{lecun2015deep}
LeCun, Y., Bengio, Y., Hinton, G.:
\newblock Deep learning.
\newblock Nature \textbf{521}(7553) (2015)

\bibitem{movshovitz2015ontological}
Movshovitz-Attias, Y., Yu, Q., Stumpe, M.C., Shet, V., Arnoud, S., Yatziv, L.:
\newblock Ontological supervision for fine grained classification of street
  view storefronts.
\newblock In: Proc. of the IEEE Conference on Computer Vision and Pattern
  Recognition. (2015)  1693--1702

\bibitem{karaoglu2017text}
Karaoglu, S., Tao, R., van Gemert, J.C., Gevers, T.:
\newblock Con-text: Text detection for fine-grained object classification.
\newblock IEEE Transactions on Image Processing \textbf{26}(8) (2017)
  3965--3980

\bibitem{bai2017integrating}
Bai, X., Yang, M., Lyu, P., Xu, Y.:
\newblock Integrating scene text and visual appearance for fine-grained image
  classification with convolutional neural networks.
\newblock arXiv preprint arXiv:1704.04613 (2017)

\bibitem{mishra2013image}
Mishra, A., Alahari, K., Jawahar, C.:
\newblock Image retrieval using textual cues.
\newblock In: Proc. of the IEEE International Conference on Computer Vision.
  (2013)  3040--3047

\bibitem{redmon2016you}
Redmon, J., Divvala, S., Girshick, R., Farhadi, A.:
\newblock You only look once: Unified, real-time object detection.
\newblock In: Proc. of the IEEE Conference on Computer Vision and Pattern
  Recognition. (2016)  779--788

\bibitem{redmon2016yolo9000}
Redmon, J., Farhadi, A.:
\newblock {YOLO9000}: better, faster, stronger.
\newblock arXiv preprint arXiv:1612.08242 (2016)

\bibitem{almazan2014word}
Almaz{\'a}n, J., Gordo, A., Forn{\'e}s, A., Valveny, E.:
\newblock Word spotting and recognition with embedded attributes.
\newblock IEEE Transactions on Pattern Analysis and Machine Intelligence
  \textbf{36}(12) (2014)  2552--2566

\bibitem{sudholt2016phocnet}
Sudholt, S., Fink, G.A.:
\newblock Phocnet: A deep convolutional neural network for word spotting in
  handwritten documents.
\newblock In: Proc. of the IEEE International Conference on Frontiers in
  Handwriting Recognition. (2016)  277--282

\bibitem{jaderberg2016reading}
Jaderberg, M., Simonyan, K., Vedaldi, A., Zisserman, A.:
\newblock Reading text in the wild with convolutional neural networks.
\newblock International Journal of Computer Vision \textbf{116}(1) (2016)
  1--20

\bibitem{gupta2016synthetic}
Gupta, A., Vedaldi, A., Zisserman, A.:
\newblock Synthetic data for text localisation in natural images.
\newblock In: Proc. of the IEEE Conference on Computer Vision and Pattern
  Recognition. (2016)  2315--2324

\bibitem{liao2017textboxes}
Liao, M., Shi, B., Bai, X., Wang, X., Liu, W.:
\newblock Textboxes: A fast text detector with a single deep neural network.
\newblock In: Proc. of the AAAI Conference on Artificial Intelligence. (2017)
  4161--4167

\bibitem{liao2018textboxes++}
Liao, M., Shi, B., Bai, X.:
\newblock Textboxes++: A single-shot oriented scene text detector.
\newblock arXiv preprint arXiv:1801.02765 (2018)

\bibitem{liu2016ssd}
Liu, W., Anguelov, D., Erhan, D., Szegedy, C., Reed, S., Fu, C.Y., Berg, A.C.:
\newblock {SSD}: Single shot multibox detector.
\newblock In: Proc. of the European Conference on Computer Vision, Springer
  (2016)  21--37

\bibitem{shi2017end}
Shi, B., Bai, X., Yao, C.:
\newblock An end-to-end trainable neural network for image-based sequence
  recognition and its application to scene text recognition.
\newblock IEEE Transactions on Pattern Analysis and Machine Intelligence
  \textbf{39}(11) (2017)

\bibitem{buvsta2017deep}
Buvsta, M., Neumann, L., Matas, J.:
\newblock Deep textspotter: An end-to-end trainable scene text localization and
  recognition framework.
\newblock In: Proc. of the IEEE International Conference on Computer Vision.
  (2017)  2204--2212

\bibitem{li2017towards}
Li, H., Wang, P., Shen, C.:
\newblock Towards end-to-end text spotting with convolutional recurrent neural
  networks.
\newblock arXiv preprint arXiv:1707.03985 (2017)

\bibitem{aldavert13}
Aldavert, D., Rusi{\~n}ol, M., Toledo, R., Llad{\'o}s, J.:
\newblock Integrating visual and textual cues for query-by-string word
  spotting.
\newblock In: Proc. of the IEEE International Conference on Document Analysis
  and Recognition. (2013)  511--515

\bibitem{ghosh2015query}
Ghosh, S.K., Valveny, E.:
\newblock Query by string word spotting based on character bi-gram indexing.
\newblock In: Proc. of the IEEE International Conference on Document Analysis
  and Recognition. (2015)  881--885

\bibitem{ren2015faster}
Ren, S., He, K., Girshick, R., Sun, J.:
\newblock Faster {R-CNN}: Towards real-time object detection with region
  proposal networks.
\newblock In: Proc. of the International Conference on Neural Information
  Processing Systems. (2015)  91--99

\bibitem{newsgroup20}
Lang, K., Mitchell, T.:
\newblock Newsgroup 20 dataset.
\newblock (1999)

\bibitem{karatzas2013icdar}
Karatzas, D., Shafait, F., Uchida, S., Iwamura, M., i~Bigorda, L.G., Mestre,
  S.R., Mas, J., Mota, D.F., Almazan, J.A., De~Las~Heras, L.P.:
\newblock {ICDAR} 2013 robust reading competition.
\newblock In: Proc. of the IEEE International Conference on Document Analysis
  and Recognition. (2013)  1484--1493

\bibitem{karatzas2015icdar}
Karatzas, D., Gomez-Bigorda, L., Nicolaou, A., Ghosh, S., Bagdanov, A.,
  Iwamura, M., Matas, J., Neumann, L., Chandrasekhar, V.R., Lu, S.,  et~al.:
\newblock {ICDAR} 2015 competition on robust reading.
\newblock In: Proc. of the IEEE International Conference on Document Analysis
  and Recognition. (2015)  1156--1160

\bibitem{wang2011end}
Wang, K., Babenko, B., Belongie, S.:
\newblock End-to-end scene text recognition.
\newblock In: Proc. of the IEEE International Conference on Computer Vision.
  (2011)  1457--1464

\bibitem{he2018single}
He, T., Tian, Z., Huang, W., Shen, C., Qiao, Y., Sun, C.:
\newblock An end-to-end textspotter with explicit alignment and attention.
\newblock In: CVPR. (2018)

\bibitem{epshtein2010detecting}
Epshtein, B., Ofek, E., Wexler, Y.:
\newblock Detecting text in natural scenes with stroke width transform.
\newblock In: Proc. of the IEEE Conference on Computer Vision and Pattern
  Recognition. (2010)  2963--2970

\bibitem{mishra2012top}
Mishra, A., Alahari, K., Jawahar, C.:
\newblock Top-down and bottom-up cues for scene text recognition.
\newblock In: Proc. of the IEEE Conference on Computer Vision and Pattern
  Recognition. (2012)  2687--2694

\bibitem{neumann2012real}
Neumann, L., Matas, J.:
\newblock Real-time scene text localization and recognition.
\newblock In: Proc. of the IEEE Conference on Computer Vision and Pattern
  Recognition. (2012)

\bibitem{ghosh2015efficient}
Ghosh, S.K., Gomez, L., Karatzas, D., Valveny, E.:
\newblock Efficient indexing for query by string text retrieval.
\newblock In: Proc. of the IEEE International Conference on Document Analysis
  and Recognition. (2015)  1236--1240

\bibitem{mishra2016understanding}
Mishra, A.:
\newblock Understanding Text in Scene Images.
\newblock PhD thesis, International Institute of Information Technology
  Hyderabad (2016)

\bibitem{gomez2017textproposals}
G{\'o}mez, L., Karatzas, D.:
\newblock Textproposals: a text-specific selective search algorithm for word
  spotting in the wild.
\newblock Pattern Recognition \textbf{70} (2017)  60--74

\bibitem{jaderberg2014synthetic}
Jaderberg, M., Simonyan, K., Vedaldi, A., Zisserman, A.:
\newblock Synthetic data and artificial neural networks for natural scene text
  recognition.
\newblock arXiv preprint arXiv:1406.2227 (2014)

\bibitem{annoy}
Bernhardsson, E.:
\newblock {ANNOY: Approximate nearest neighbors in C++/Python optimized for
  memory usage and loading/saving to disk. (2013)}

\bibitem{malkov2016efficient}
Malkov, Y.A., Yashunin, D.:
\newblock Efficient and robust approximate nearest neighbor search using
  hierarchical navigable small world graphs.
\newblock arXiv:1603.09320 (2016)

\bibitem{andoni2015practical}
Andoni, A., Indyk, P., Laarhoven, T., Razenshteyn, I., Schmidt, L.:
\newblock Practical and optimal {LSH} for angular distance.
\newblock In: NIPS. (2015)

\end{thebibliography}

\end{document}